\documentclass{article}
\usepackage{kr}

\usepackage{times}
\usepackage{soul}
\usepackage{url}
\usepackage[hidelinks]{hyperref}
\usepackage[utf8]{inputenc}
\usepackage[small]{caption}
\usepackage{graphicx}
\usepackage{amsmath}
\usepackage{amsthm}
\usepackage{booktabs}
\usepackage{algorithm}
\usepackage{algorithmic}
\usepackage{mathtools}

\usepackage{amsmath}
\usepackage{amssymb}
\usepackage{xcolor}
\usepackage{bbm}
\usepackage{tikz}
\usepackage{subfig}

\usepackage{marvosym}
\usepackage{tikzsymbols}

\usetikzlibrary{arrows,automata,positioning,calc}

\newtheorem{example}{Example}
\newtheorem{theorem}{Theorem}

\newtheorem{definition}{Definition}

\newtheorem{lemma}{Lemma}

\newenvironment{hproof}{%
  \proof}{\endproof}

\newcommand{\pvec}[1]{\vec{#1}\mkern2mu\vphantom{#1}} 

\newcommand{\cn}{{\mathbb{C}}}
\newcommand{\band}{\bigwedge}
\newcommand{\all}{\forall}

\newcommand{\A}{{\cal A}}

\newcommand{\C}{{\cal C}}
\newcommand{\D}{{\cal D}}

\newcommand{\lang}{\mathcal{DS}_p}

\newcommand{\N}{{\cal N}}

\newcommand{\calP}{{\cal P}}

\newcommand{\R}{{\cal R}}
\newcommand{\T}{{\cal T}}
\newcommand{\U}{{\cal U}}
\newcommand{\V}{{\cal V}}
\newcommand{\W}{{\cal W}}
\newcommand{\wo}{\boldsymbol{\mathsf{w}}}

\newcommand{\Z}{{\cal Z}}
\newcommand{\pp}{{\varrho}}

\newcommand{\pe}{\delta}
\newcommand{\lan}{\langle}
\newcommand{\ran}{\rangle}

\newcommand{\union}{\cup}
\newcommand{\ls}{l^*}

\DeclareMathAlphabet{\mathitbf}{OML}{cmm}{b}{it}    
\newcommand{\bel}{\mathitbf{B}}
\renewcommand{\exp}{\mathitbf{Exp}}
\newcommand{\conf}{\mathitbf{Conf}}

\newcommand{\oknow}{\mathitbf{O}}

\newcommand{\MDP}{\mathsf{M}}
\newcommand{\trans}{\xrightarrow{e}}
\newcommand{\etrans}{\xhookrightarrow{e}}

\newcommand{\Fin}{\mathrm{Fin}(e)}
\newcommand{\Fail}{\mathrm{Fail}(e)}
\newcommand{\e}{\mathfrak{\epsilon}}
\newcommand{\f}{\mathfrak{f}}

\newcommand{\dtgolog}{\textsc{DT-Golog}} 
\newcommand{\golog}{\textsc{Golog}} 

\newcommand{\normal}{\textsc{Norm}} 

\newcommand{\equal}{\textsc{Eq}}
\newcommand{\bound}{\textsc{Bnd}}

\newcommand{\eq}{\equal}
\newcommand{\KB}{\mbox{KB}}

\newcommand{\true}{{\mbox{\footnotesize {\sc true}}}}
\newcommand{\false}{\mbox{\footnotesize {\sc false}}}

\definecolor{darkpastelgreen}{rgb}{0.01, 0.75, 0.24}

\title{On the Verification of Belief Programs}

\author{%
Daxin Liu\and
Gerhard Lakemeyer\\
\affiliations
RWTH Aachen University\\
\emails
\{liu, gerhard\}@kbsg.rwth-aachen.de
}

\begin{document}

\maketitle

\begin{abstract}

In a recent paper, Belle and Levesque proposed a framework for a type of program called belief programs, a probabilistic extension of GOLOG programs where every action and sensing result could be noisy and every test condition refers to the agent's subjective beliefs. Inherited from GOLOG programs, the action-centered feature makes belief programs fairly suitable for high-level robot control under uncertainty. An important step before deploying such a program is to verify whether it satisfies properties as desired. At least two problems exist in doing verification: how to formally specify properties of a program and what is the complexity of verification.
In this paper, we propose a formalism for belief programs based on a modal logic of actions and beliefs. Among other things, this allows us to express PCTL-like temporal properties smoothly. Besides, we investigate the decidability and undecidability for the verification problem of belief programs.

\end{abstract}

\section{Introduction}\label{sec:introduction}

The $\golog$ \cite{levesque1997golog} family of agent programming language has been proven to be a powerful means to express high-level agent behavior. Combining $\golog$ with probabilistic reasoning, Belle and Levesque \shortcite{belle2015allegro} proposed an extension called belief programs, where every action and sensing result could be noisy. 
Along with the feature that test conditions refer to the agent's subjective beliefs, belief programs are fairly suitable for robot control in an uncertain environment.

\begin{figure}[t]
  \centering
   \includegraphics[width=.3\textwidth]{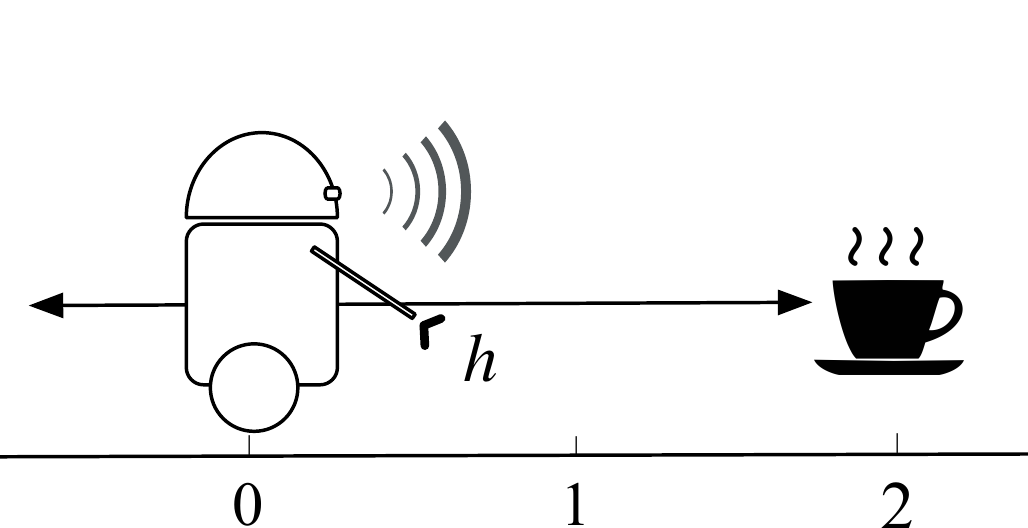}
  \caption{A coffee searching robot.}
  \label{fig:coffeesearch}
\end{figure}

For safety and economic reasons, verifying such a program to ensure that it meets certain properties as desired before deployment is essential and desirable. As an illustrative example, consider a robot searching for coffee in a one-dimensional world as in Fig~\ref{fig:coffeesearch}. Initially, the horizontal position $h$ of the robot is at 0 and the coffee is at 2. Additionally, the robot has a knowledge base about its own location (usually a belief distribution, e.g. a uniform distribution among two points $\{0,1\}$). The robot might perform noisy sensing \(sencfe\) to detect whether its current location has the coffee or not and an action \(east(1)\) to move 1 unit east. 
A possible belief program is given in Table.~\ref{tab:bpexample}. The robot continuously uses its sensor to detect whether its current location has the coffee or not (line 2-3). When it is confident enough,\footnote{The agent's confidence $\conf(h,n)$ of a random variable $h$ wrt a number $n$ is defined as its belief that $h$ is somewhere in the interval $[\exp(h)-n,\exp(h)+n]$, here $\exp(h)$ is the expectation of $h$. 
} it tries to move 1 unit east (line 5). If it still does not fully believe it reached the coffee, i.e. position at 2 (line 1), it repeats the above process. The program is an online program as its execution depends on the outcome of sensing.

\begin{table}
    \centering
    \begin{tabular}{|l|}
    \hline
    1 \textbf{while} ~\(\bel(h=2) < 1\) ~\textbf{do}\\
    2 \quad \textbf{while} ~\(\conf(h,0.5) \le 0.5\) ~\textbf{do} \\
    3    \quad\quad   \emph{sencfe};\\
    4    \quad \textbf{endWhile}\\  
    5\quad \emph{east(1)};\\
    6 \textbf{endWhile}\\
    \hline
    \end{tabular}
    \caption{A online belief program for the robot.}
    \label{tab:bpexample}
\end{table}

Some interesting properties of the program are:

\begin{enumerate}
    \item \textbf{P1}: whether the probability that within 2 steps of the program the robot \emph{believes} it reached the coffee with certainty is higher than 0.05;
    
    \item \textbf{P2}: whether it is almost certain that \emph{eventually} the robot \emph{believes} it reached the coffee with certainty.
\end{enumerate}


Often, the above program properties are specified by temporal formulas via Probabilistic Computational Tree Logic (PCTL) in model checking. Obtaining the answers is non-trivial as the answers depend on both the physical world (like the robot's position and action models of actuators and sensors) and the robot's epistemic state (like the robot's beliefs about its position and action models). There are at least two questions in verifying belief programs: 1. how can we formally specify temporal properties as above; 2. what is the complexity of the verification problem?

The semantics of belief programs proposed by Belle and Levesque \shortcite{belle2015allegro} is based on the well-known BHL logic \cite{bacchus:bhl} that combines \emph{situation calculus} and probabilistic reasoning in a purely axiomatic fashion. 
While verification has been studied in this fashion in the non-probabilistic case~\cite{de2019non}, it is somewhat cumbersome as it relies heavily on the use of second-order logic and the $\mu$-calculus.
For instance, consider a domain where the robot is programmed to serve coffee for guests on request \cite{classen2013planning}. An interesting property of the program is whether \emph{every request will eventually be served}. Such a property is then expressed  as follows:
\[
\begin{aligned}
 (\all x,\delta,s) Trans^*(\delta_0,S_0,\delta, do(requestCoffee(x),s)) \\ \supset EventuallyServed(x,\delta,do(requestCoffee(x),s))
\end{aligned}
\]

\noindent where $Trans^*$ refers to the transitive closure of the $Trans$ predicate (a predicate axiomatically defining the transitions among program configurations) and $EventuallyServed$ is defined by
\[
\begin{aligned}
Even&tuallyServed(x,\delta_1,s_1)  ::=\\
 & \mu_{P,\delta,s} \{[(\exists s'')s=do(selectRequest(x),s'')] \vee\\
&  \ \quad\quad[((\exists \delta',s') Trans(\delta,s,\delta',s')) \land \\
&(\all \delta',s') Trans(\delta,s,\delta',s')\supset P(\delta',s')]\}(\delta_1,s_1).
\end{aligned}
\]

\noindent Here the notion $\mu_{P,\delta,s}$ denotes a least fixpoint according to the  formula 
$(\all \pvec{x}\,)\{\mu_{P,\pvec{y}}\Phi(P,\pvec{y}\,)(\pvec{x}\,)
\equiv[(\all P)[(\all $ $ y)\Phi(P,\pvec{y}\,) \supset P(\pvec{y}\,)]\supset P(\pvec{x}\,)]
\}$
. We do not go into more details here but refer interested readers to \cite{classen2013planning}.




In this paper, we propose a new semantics for belief programs based on the logic \(\lang\)~\cite{liu:dsp}, a modal version of the BHL logic with a possible-world semantics. Such a modal formalism makes it smoother than axiomatic approaches to express temporal properties like \emph{eventually}  and \emph{globally}  by using the the usual modals $\mathbf{F}$ and $\mathbf{G}$ in temporal logic. 
Subsequently, we study the boundary of decidability of the verification problem. As it turns out, the result is strongly negative. However, we also investigate a case where the problem is decidable.

The rest of the paper is organized as follows. In section \ref{sec:foundation}, we introduce the logic \(\lang\). Subsequently, we present the proposed semantics and specification of temporal properties for belief programs in section~\ref{sec:framework}. In section \ref{sec:undecidability}, we study the boundary of decidability of the verification problem in a specific dimension. Section \ref{sec:decidability} considers a special case where the problem is decidable. In section \ref{sec:relatedwork} and \ref{sec:conclusion}, we review related work and conclude.

\section{Logical Foundation}
\label{sec:foundation}

\subsection{The Logic \texorpdfstring{\(\lang\)}{DS-p}} 

The logic \(\lang\) is a modal variant of the epistemic situation calculus.
There are two sorts: {\it object} and {\it action}. Implicitly, we assume that {\it number} is a sub-sort of object and refers to the {\it computable numbers} $\cn$.\footnote{We use the computable numbers as they are still enumerable and allow us to refer to certain real numbers such as $\sqrt{2}$ and Euler's number $e$.}

\subsubsection{The Language} 
We use \(\lang\)'s first-order fragment with equality. The logic features a countable set of so-called \emph{standard names} $\N$, which are isomorphic with a fixed universe of discourse. Roughly, this amounts to having an infinite domain closure axiom together with the unique name assumption. $\N = \N_{O} \union \N_{A}$ where $\N_{O}$ and $\N_{A}$ are standard object names and standard action names, respectively.
Function symbols are divided into \emph{fluent function symbols} and \emph{rigid function symbols}. 
For simplicity, all action functions are rigid and we do not include predicate symbols. Fluents vary as the result of actions, yet denotations of rigid functions are fixed. The language includes modal operators $\bel$ and $\oknow$ for \emph{degrees of belief} and \emph{only-believing}, respectively.
Finally, there are two special fluent functions: a function $l(a)$ specifies action $a$'s likelihood and a binary function $oi$ encodes the \emph{observational indistinguishability} among actions. The idea is that in an uncertain setting, instead of saying an action might have non-deterministic effects, we say the action is stochastic and has non-deterministic alternatives, which are observationally indistinguishable by the agent and each of which has deterministic effects.


The terms of the language are formed in the usual way from variables, standard names and function symbols.
A term is said to be \emph{rigid} if it does not mention fluents. \emph{Ground terms} are terms without variables. \emph{ Primitive terms} are terms of the form \(f(n_1,\ldots, n_k) \), where $f$ is a function symbol and \(n_i\) are standard  object names. We denote the sets of primitive terms of sort object and action as $\calP_{O} $ and $ \calP_{A}$, respectively. While standard  object names are syntactically like constants, we require that standard action  names are all the primitive action terms, i.e. \(\N_{A} =  \calP_{A} \). For example, the sensing action $sencfe(1)$, where the robot receives a positive signal, is considered as a standard action name.
Furthermore, $\Z$ refers to the set of all finite sequences of standard action  names, including the empty sequence $\lan\ran$. We reserve standard names \( \top, \bot\)  in \( \N_{O} \) for truth values (to simulate predicates).



Atomic formulas are expressions of the form $t_1=t_2$ for terms $t_1,t_2$. Arbitrary formulas are formed with the usual logical operators $\neg,\land$, the quantifier $\all$, and modal operators $[t_a]$, where $t_a$ is an action term, $\Box$,
\( \bel(\alpha_1\colon r_1) \) and $\oknow(\alpha_1: r_1,\ldots, \alpha_k: r_k)$, where the $\alpha_i$ are formulas and the $r_i$ rigid terms of sort number.

\( [t_a]\alpha \) should be read as  ``$\alpha$ holds after action $t_a$," \( \Box \alpha \) as ``$\alpha$ holds after any sequence of actions,'' \( \bel(\alpha\colon r) \) as ``\( \alpha \) is believed with a probability \( r\)''.
\( \oknow(\alpha_1: r_1,\ldots, \alpha_k: r_k) \) may be read as ``the \(\alpha_i\) with a probability \(r_i\) are all that is believed''. Similarly, \( \oknow\alpha \) means ``\(\alpha\) is only known'' and is an abbreviation for \( \oknow(\alpha \colon 1) \).  For action sequence \( z = t_1\cdots t_k, \) we  write \( [z]\alpha \) to mean \( [t_1]\cdots[t_k]\alpha\). $\alpha^x_t$ is the formula obtained by substituting all free occurrences of $x$ in $\alpha$ by $t$. As usual, we treat $\alpha \vee \beta$, $\alpha \supset \beta$, $\alpha \equiv \beta$, and $\exists v. \alpha$ as abbreviations.

 A \emph{sentence} is a formula without free variables.  We use \( \true \) as an abbreviation for \( \forall x(x=x), \) and \( \false \) for its negation.
 A formula with no $\Box$ is called \emph{bounded}. 
 A formula with no $\Box$ or $[t_a]$ is called \emph{static}. A formula with no $\bel$ or $\oknow$ is called \emph{objective}. A formula with no fluent, $\Box$ or $[t_a]$ outside $\bel$ or $\oknow$  is called \emph{subjective}. A formula with no $\bel$, $\oknow$, $\Box$, $[t_a]$, $l$, $oi$ is called  a \emph{fluent formula}. A fluent formula without fluent functions is called a \emph{rigid formula}.

\subsubsection{The Semantics} The semantics is given in terms of possible worlds. A \textbf{world} \(w\) is a mapping from the primitive terms ($\calP_{O}\union \calP_{A}$) and \(\Z\) to $\N$ of the right sort, satisfying \emph{rigidity} and \emph{arithmetical correctness}.\footnote{ \emph{Rigidity:} If \(t\) is rigid, then for all \((w,z),(w',z')\), \(w[t,z]=w'[t,z']\). \emph{Arithmetical Correctness:} 
Any arithmetical expression is rigid and has its standard value.} We denote the set of all such worlds as $\W$. 
Given \(w\in \W\), \(z \in \Z\), and a ground term \(t\), we define \(|t|^{z}_{w}\) (the denotation for \(t\) given \(w,z\)) by: 

\begin{enumerate}
    \item If $t \in \N$, then $|t|^{z}_{w} = t$;
    \item $|f(t_1,\ldots, t_k)|^{z}_{w} = w[f(|t_1|^{z}_{w}, \ldots, |t_k|^{z}_{w}),z]$.
\end{enumerate} 


For a rigid ground term \(t\), we use $|t|$ instead of $|t|^{z}_{w}$.
We will require that 
\(l(a)\)  is of sort number,  and \(oi(a,a')\) only takes values \(\top\) or \(\bot \), and \( oi \) is an equivalence relation (reflexive, symmetric, and transitive). Intuitively, $l(a)$ denotes the likelihood of action $a$, while \(oi(a,a')\) means $a$ and $a'$ are mutual alternatives. 
In the example of Fig.~\ref{fig:coffeesearch},
the robot might perform a stochastic action $east(x,y)$, where $x$ is its intended moving distance and $y$ is the actual outcome selected by nature.
Then, $oi(east(1,0),east(1,1))$ says that nature can non-deterministically select 0 or 1 as a result for the intended value 1. 

A \textbf{distribution} \( d\) is a mapping from \( \W \) to \( \mathbb{R}^{\ge 0} \) and an \textbf{epistemic state} \( e \) is any set of distributions. By a \textbf{model}, we mean a triple ($e,w,z$).

To account for \(\bel\) and $\oknow$ after actions, we need to extend the fluents $l$, $oi$ from actions to action sequences:

\begin{definition} Given a world \(w\), we define:
 \begin{enumerate}
	  \item \(\ls \colon \W \times \Z \mapsto \mathbb{R}^{\ge 0}\) as  
	  		   
	  		   \( \ls(w,\lan \ran) = 1 \); 
                	 
                \( \ls(w, z\cdot a) = \ls(w,z) \times n \) where \( w[l(a),z] = n\).
	      
	 \item \(z \sim_w z'\) as
		    
            \( \lan \ran \sim_w z' \)  iff \( z' = \lan\ran\); 
            
            \( z \cdot a \sim_w z' \) iff \( z' = z^* \cdot a^* \), \( z \sim_w z^* \), \( w[oi(a,a^*),z] = \top \).
\end{enumerate}	
\end{definition}

To obtain a well-defined sum over uncountably many worlds, some conditions are used for $\bel$ and $\oknow$:

\begin{definition} We define \( \bound , \equal, \normal \) for any distribution \( d \) and any set \( \V = \{ (w_1, z_1), (w_2, z_2), \ldots \} \) as follows:

\begin{enumerate}
    \item \( \bound(d, \V, r) \) iff  \( \neg\exists  k, (w_1, z_1), \ldots, (w_k, z_k) \in \V \) such that $
	    \sum_{i=1}^k d(w_i) \times \ls(w_i, z_i)  >r. $
	\item \( \equal(d, \V, r) \) iff \( \bound(d, \V, r) \) and there is no \( r' <r\) such that \( \bound(d, \V, r') \) holds. 

	\item for any \( \U \subseteq \V, \) \( \normal(d, \U, \V, r) \) iff \( \exists b \neq 0 \) such that \( \equal(d,\U, b\times r) \) and \( \equal(d,\V, b). \)
\end{enumerate}
\end{definition}

Intuitively, given \(\normal(d, \U, \V, r)\), \(r\) can be viewed as the \emph{normalized} sum of the weights of worlds in \(\U\) wrt $d$ in relation to \(\V\). Here \( \equal(d,\V,b)\) expresses that the weight of the worlds wrt $d$ in \(\V\) is \(b\), and finally  \( \bound(d,\V,b)\) ensures the weights of worlds in \(\V\) is bounded by \(b\). In essence, even if \(\W\) is uncountable, the condition \(\normal\) ensures $d$ is in fact discrete, i.e. only countably many worlds have non-zero weight wrt $d$~\cite{belle:obl}.

The truth of sentences in \( \lang \) is defined as: 

\begin{itemize}
	\item \( e,w,z \models t_1 = t_2 \) iff \( |t_1|^z_{w}~\textrm{ and }~ |t_2|^z_{w}\) are identical;
	
	\item \( e,w,z\models \neg\alpha \) iff \( e,w,z\not\models \alpha \); 

	\item \( e,w,z \models \alpha\land\beta \) iff \( e,w,z\models \alpha \) and \( e,w,z \models \beta \); 

	\item \( e,w,z\models \forall x.\alpha \) iff \( e,w,z \models \alpha^x_n \) for every standard name \(n\) of the right sort;
	
	\item \( e,w,z\models [t_a]\alpha \) iff \( e,w,z\cdot n\models \alpha \) and \(n=|t_a|^z_w\);
	\item \( e,w,z\models \Box \alpha \) iff \( e,w,z\cdot z'\models \alpha \) for all \( z' \in \Z \).
\end{itemize}

To prepare for the semantics of epistemic operators, let \( \W^{e,z}_\alpha = \{ (w', z') \mid z' \sim_{w'} z, ~\textrm{and}~ e, w', \lan \ran \models [z'] \alpha \} \). If \(z = \lan \ran\), we ignore $z$ and write \(\W^{e}_\alpha\). If the context is clear, we write \( \W_\alpha\). Intuitively, \( \W_\alpha\) is the set of alternatives (world and action sequence pairs) of $z$ that might result in $\alpha$.  A distribution \(d\) is \textbf{regular} iff 
\(\eq(d,\W_\true^{\{d\}}, n)\) for some $n \in \mathbb{R}^{>0}$. We denote the set of all regular distributions as $\D$.

\begin{definition}
Given $w\in \W, d \in \D, z\in \Z$, we define
\begin{itemize}
    \item $w_z$ as a world such that for all primitive terms $t$ and $z' \in \Z$, $w_z[t,z'] = w[t,z\cdot z']$;
    
    \item $d_{z}$ a mapping such that for all $w \in \W$,
    
    \(d_{z}(w)=\sum_{\{w' : d(w')>0\}} \sum_{\{z' : z'\sim_{w'}z\textrm{,}~w'_{z'} =w\}} d(w') \times \ls(w',z')\). 
\end{itemize}
\end{definition}

$w_z$ is called the \textbf{progressed world} of $w$ while $d_z$ is called the \textbf{progressed distribution} wrt $z$. 
A remark is that the $d_z$ might not be regular for a regular $d$. For example, if the likelihood of a ground sensing action $t_{sen}$ is zero in all worlds with non-zero weights, then $\eq(d_{t_{sen}}, \W^{d_{t_{sen}}}_\true, 0)$. Hence we define:
\begin{definition}
\label{def:comp}
A distribution \(d\) is \textbf{compatible} with action sequence \(z\), $d\sim_{comp}z$ iff $d_{z} \in \D$;
given an epistemic state $e$, the set $e_z = cl(\{d_z | d\in e\cap \D, d\sim_{comp}z\})$ is called the \textbf{progressed epistemic state} of $e$ wrt $z$, here $cl(\cdot)$ is a closure operator.\footnote{More precisely, $cl(\cdot)$ is the closure operator of the metric space $(\D,\rho)$ where $\rho$ is a distance function defined as $\rho(d,d') = \sum_{ w \in \W} |d(w)-d'(w)|$ for $d,d' \in \D$. The closure operator is important to ensure a correct semantic of progression in $\lang$ as Liu and Feng \shortcite{liu:dsp}
shows that the set of discrete distributions that satisfies a given belief is a closed set in $(\D,\rho)$.}
\end{definition}

Intuitively, $d\sim_{comp} z$ ensures $z$ has non-zero likelihood in at least one world whose weight is non-zero in $d$. As a consequence, $d\sim_{comp} \lan \ran$ iff $d \in \D$. Note that the progressed epistemic state of $e$ is only about its regular subset $e\cap \D$ and $e_z \subseteq \D$, therefore $e \neq e_{\lan \ran}$ in general.

The truth of \(\bel\) and $\oknow$ is given by:
\begin{itemize}
    \item \(e,w, z  \models \bel(\alpha \colon  r)\)  iff \(\all d \in e_z\), \\ \(\normal(d
,\W_\alpha^{\{d\}}, \W_\true^{\{d\}}, n)\) for $n\in \cn$ and \(n=|r|\);

    \item \(e,w, z  \models \oknow(\alpha_1 \colon r_1,\ldots, \alpha_k \colon r_k  ) \)  iff \(\all d \), \(d \in e_{z}\) iff for all \(1 \le i \le k\), \(\normal(d,\W_{\alpha_i}^{\{d\}}, \W_\true^{\{d\}}, n_i)\) for $n_i\in \cn$, and \(n_i=|r_i|\);
\end{itemize}

 For any sentence \( \alpha \), we   write \( e,w\models \alpha \) instead of \( e,w, \lan\ran \models \alpha \). When \( \Sigma  \) is a set of sentences and \( \alpha \) is a sentence, we write \( \Sigma \models \alpha \) (read: \( \Sigma \) logically entails \( \alpha \)) to mean that for every set of regular distributions \( e \) and \( w \), if \( e,w\models \alpha' \) for every \( \alpha'\in\Sigma \), then \( e,w\models\alpha \). 
 We say that $\alpha$ is valid \( (\models\alpha) \) if $\{\}\models\alpha$. Satisfiability is then defined in the usual way.
 If \( \alpha \) is an objective formula, we write \( w \models \alpha \) instead of \(e,w \models \alpha \). Similarly, we write \( e \models \alpha \) instead of \( e,w \models \alpha \) if $\alpha$ is subjective.

\subsection{Basic Action Theories and Projection}

Besides the usual $+,\times$, it is desirable to include some usual mathematical functions as logical terms. We achieve this by axioms. We call these axioms  \emph{definitional axioms},\footnote{In the rest of the paper, whenever we write logical entailment $\Sigma \models \alpha$, we implicitly mean $\Sigma \union \Delta \models \alpha$, where $\Delta$ is the set of all definitional axioms of functions involved in $\Sigma$ and $\alpha$.} such functions as \emph{definitional functions}, and terms constructed by definitional functions as \emph{definitional terms}. E.g. the following axiom specifies the uniform distribution $\mathbf{U}_{\{0,1\}}$.
\begin{equation}
    \label{exp:uniform}
    \begin{aligned}
        \all v .\all u. \mathbf{U}_{\{0,1\}}(u)= v \equiv (u=0 \vee u=1) \land v=0.5 \\
    \vee \neg (u=0 \vee u=1) \land v=0
    \end{aligned}
\end{equation}

\subsubsection*{Basic Action Theories}
BATs were first introduced by Rei{\-}ter~\shortcite{reiter:knowledge} to describe the dynamics of an application domain. Given a finite set of fluents $\mathcal{H}$, a BAT $\Sigma$ over $\mathcal{H}$ consists of the union of the following sets:
\begin{itemize}
\item $\Sigma_{post}$: A set of \emph{successor state axioms} (SSAs), one for each fluent $h$ in $\mathcal{H}$, of the form $\Box [a]h(\pvec{p})= u \equiv \gamma_h$\footnote{Free variables are implicitly universally quantified from the outside. The $\Box$ modality  has lower syntactic precedence than the connectives,
and $[\cdot]$ has the highest priority.}
to characterize action effects, also providing a solution to the frame problem \cite{reiter:knowledge}. Here $\gamma_h$ is a fluent formula with free variables $\vec{p},u$ and it is functional in $u$, 

\item $\Sigma_{oi}$: A single axiom of the form  $\Box oi(a,a') =\top \equiv \psi$ to represent the observational indistinguishability relation among actions. Here $\psi$ is a rigid formula.\footnote{The rigidity here is crucial for properties like introspection and regression, see \cite{liu:ijcai}.}

\item $\Sigma_{l}$: A single likelihood axiom (LA) of the form \( \Box l(a) = \mathcal{L}(a) \), here $\mathcal{L}(a)$ is a definitional term with $a$ free.

\end{itemize}


Besides BATs, we need to specify what holds initially. This is achieved by a set of fluent sentences $\Sigma_0$.
By \textbf{belief distribution}, we mean the joint distribution of a finite set of random variables. Formally, assuming all fluents in $\mathcal{H}$ are \textbf{nullary},\footnote{Allowing fluents with arguments would result in joint distribution over infinitely many random variables, which is generally problematic in probability theory~\cite{belle2018reasoning}.} \(\mathcal{H}=\{h_1,\ldots, h_m\}\),  a belief distribution \(\bel^f\) of $\mathcal{H}$ is a formula of the form 
\(\all \pvec{u}. \bel(\pvec{h} = \pvec{u}\colon f(\pvec{u}))  
\), where \(\pvec{u}\) is a set of variables, $\pvec{h} = \pvec{u}$ stands for $\band h_i = u_i$, and \(f\) is a definitional function of sort number with free variables \(\pvec{u}\).
Finally, by a knowledge base ($\KB$), we mean a sentence of the form $\oknow(\bel^{{f}} \land \Sigma)$. Note that the BAT of the actual world is not necessarily the same as the BAT believed by the agent.

\begin{example}
\label{example:bat}
The following is a BAT $\Sigma$ for our coffee robot:
\(
\begin{aligned}
&\Box[a]h=u \equiv \exists x,y.a=east(x,y) \land u=h+y \\
    &\vee \all x,y.a\neq east(x,y) \land h=u\\
 &\Box oi(a,a')=\top \equiv  \exists x,y,y'. a = east(x,y) \\
 &\land a' = east(x,y') 
 \vee \exists y.a =sencfe(y)\land a' = a\\
   &\Box l(a)= \mathcal{L}(a) \textrm{~with}
\end{aligned}
\)

\(
\begin{aligned}
    \mathcal{L}(a) = \left \{\begin{array}{cc}
      \mathbf{U}_{\{x,x-1\}}(y)  &  \exists x,y. a=east(x,y)\\
        \theta_{noisy}(h,y)  & \exists y.a=sencfe(y)
    \end{array} \right.
\end{aligned}
\)

\noindent where $\theta_{noisy}(x,y)$ is defined as~\footnote{Here, ``$\in$'' should be understood as a finite disjunction. For
readability, we write the definitional functions in this form, they
should be understood as logical formulas as Eq.~(\ref{exp:uniform}).
}

\begin{equation}
\label{eq:sigmalnoisy}
\resizebox{1\linewidth}{!}{
\(
    \theta_{noisy}(x,y) = \left\{ \begin{array}{cc}
        \theta(x) & y=1 \\
        1-\theta(x) & y=0
    \end{array}\right.
~
    \theta(x) = \left \{\begin{array}{cc}
       0.1  &  x \in \{1,3\}\\
        0.8  & x=2 \\
        0 & o.w.
    \end{array} \right.
\)
}
\end{equation} 

\noindent A possible initial state axiom could be $\Sigma_0=\{h\le 0\}$ and a possible $\KB$ is $\oknow(\bel^{{f}} \land \Sigma')$ where
$\Sigma'$ is exactly the same as $\Sigma$ with $\theta_{noisy}(x,y)$ in Eq.~(\ref{eq:sigmalnoisy}) replaced by $\theta_{acc}(x,y)$
\resizebox{1\linewidth}{!}{
\(
    \theta_{acc}(x,y) = \left\{ \begin{array}{cc}
        \theta'(x) & y=1 \\
        1-\theta'(x) & y=0
    \end{array}\right.
~
\begin{aligned}
    \theta'(x) = \left \{\begin{array}{cc}
       0  &  x\neq2\\
        1  & x=2
    \end{array} \right.
\end{aligned}
\)}
and $f(u) = \mathbf{U}_{\{0,1\}}(u)$
\end{example}

In English, the robot's position \(h\) can only be affected by \(east(x,y)\) and the value is determined by nature's choice \(y\), not the intended value \(x\); the exact distance moved is unobservable to the agent; for the stochastic action $east(x,y)$, with the half-half likelihood the exact distance $y$ moved  equals to or is 1 unit less than the intended value $x$ ($\mathbf{U}_{\{x,x-1\}}(y)$);
sensing action $sencfe(y)$ is noisy and there are only two possible outcomes $y \in \{0,1\}$ ($1$ for coffee-sensed and $0$ otherwise);
additionally, the likelihood of $sencfe(y)$ depends on the robot's position $h$ ($\theta_{noisy}(h,y)$): when the robot is at 2 ($x=2$) where the coffee is located, with a high likelihood (0.8), sensing returns 1 and when the robot is 1 unit away from the position 2 ($x \in \{1,3\}$), with a low likelihood (0.1), sensing returns 1.
Initially, the robot is at a certain non-positive position and it believes its position distributes uniformly among \{0,1\}. Furthermore, although its sensor is noisy, it believes the sensor is accurate ($\theta_{acc}(x,y)$).

\subsubsection{Projection by Progression} Projection in general is to decide what holds after actions. Progression is a solution to projection and the idea is to change the initial state according to the effects of actions and then evaluate queries against the updated state.
Lin and Reiter
\shortcite{lin1997progress} showed that 
progression is \emph{only} second order definable in general. However, Liu and Feng \shortcite{liu:dsp} showed that if all fluents are nullary, for the objective fragment, progression is first-order definable. Let $Pro(\Sigma_0,\Sigma,t)$ be the FO progression of $\Sigma_0$ wrt $\Sigma$ and action term $t$ ($Pro(\Sigma_0,t)$ for short). They also showed that the progression of a $\KB$  $\oknow(\bel^f\land \Sigma)$ wrt to a stochastic action $t$, denoted by $Pro(\oknow(\bel^f\land \Sigma),t)$, is another \(\KB\) $\oknow(\bel^{f'} \land \Sigma)$ with a belief distribution $\bel^{f'}$ and

\[f'(\pvec{u}) = \sum_{\pvec{u}' \in (\N_O)^m } f(\pvec{u}') \sum_{a \in \N_A}   \mathcal{L}(a)^{\pvec{h}}_{\pvec{u}'} \times \mathbb{I}(\pvec{u},\pvec{u}',a,t)\]

\noindent where $\mathbb{I}$ is a definitional function given by

 \(
\mathbb{I}(\pvec{u},\pvec{u}',a,t) = \left \{
\begin{array}{cc}
     1 &  Pro(\pvec{h}=\pvec{u}', a )^{\pvec{h}}_{\pvec{u}} \land (\psi)_t^{a'}\\
    0 & o.w.
\end{array}
\right .\)

\noindent Here $\psi$ is the RHS of $\Sigma_{oi}$. If $t$ is a sensing action, then $f'$ is given by  
\(f'(\pvec{u}) = \frac{1}{\eta} f (\pvec{u}) \times  \mathcal{L}(t)^{\pvec{h}}_{\pvec{u}}\), and $\eta$ is a normalizer as \( \eta =  \sum_{\pvec{u}' \in (\N_O)^m } f(\pvec{u}') \times  \mathcal{L}(t)^{\pvec{h}}_{\pvec{u}'}\). 

\begin{example}
\label{example:progression}

Let $\oknow(\bel^f \land \Sigma')$ be as in  Example~\ref{example:bat}, then its progression wrt the stochastic action $east(1,1)$ is $\oknow(\bel^{f'} \land \Sigma')$, the progression of $\oknow(\bel^{f'} \land \Sigma')$ wrt the sensing action $sencfe(1)$ is $\oknow(\bel^{f''} \land \Sigma')$ where $f'$ and $f''$ are given by:

\noindent\(
    f'(u) = \left\{ \begin{array}{cc}
        \frac{1}{4} & u \in \{0,2\}\\
        \frac{1}{2} & u=1\\
        0 & o.w.
    \end{array}\right.
\textrm{and}
~
\begin{aligned}
    f''(u) = \left \{\begin{array}{cc}
       1  &  u=2\\
     0  & o.w.
    \end{array} \right.
\end{aligned}
\)
\end{example}


\subsubsection{Avoiding Infinite Summation} A notable point above is that progression requires infinite summation. $\lang$ treats summation as a rigid logical term just like $+,\times$ and disregards the computational issues therein. Nevertheless, to ensure decidability of the logic, one needs to avoid infinite summation.


Consequently, we have the following restrictions. Firstly, we assume that only two types of action symbol are used: stochastic actions $sa_1,\ldots, sa_k$ and sensing $sen_1,\ldots, sen_{k'}$.  Moreover, parameters of stochastic action $sa(\pvec{x},\pvec{y}\,)$ are divided into two parts, where \(\pvec{x}\) is a set of \emph{controllable} and \emph{observable}  parameters and \(\pvec{y}\) is a set of \emph{uncontrollable} and \emph{unobservable} parameters. Parameters of sensing $sen(\pvec{y}\,)$ are all \emph{observable} yet \emph{uncontrollable} by the agent. Additionally, we require:
 
 
     
\begin{enumerate}
    \item $\psi$ in $\Sigma_{oi}$  has the form $\psi \equiv \psi_{sa} \vee \psi_{sen}$  with \(\psi_{sa} \equiv \bigvee_i \exists \pvec{x}. \exists \pvec{y}.\exists \pvec{y}'. a = sa_i(\pvec{x},\pvec{y}\,) \land a' = sa_i(\pvec{x},\pvec{y}') \) and  \(\psi_{sen} \equiv \bigvee_j \exists \pvec{y}. a = sen_j(\pvec{y}\,) \land a=a' \);
    
     \item $\Sigma_{l}$  is of the form
     $
     \Box l(a)=v \equiv \bigvee_i \exists \pvec{x},\pvec{y}.a=sa_i(\pvec{x},\pvec{y}\,) \land v = \mathcal{L}_{sa_i}(\pvec{x},\pvec{y}\,)$ $
     \vee \bigvee_i \exists \pvec{y}.a=sen_i(\pvec{y}\,) \land v = \mathcal{L}_{sen_i}(\pvec{y}\,)
     $     
    where $\mathcal{L}_{sa_i}$ and $\mathcal{L}_{sen_i}$ are given by: (free variables are implicitly universally quantified from the outside)
$$
\resizebox{1\linewidth}{!}{
\(
\begin{aligned}
&\mathcal{L}_{sa_i}(\pvec{x},\pvec{y}\,)=v \equiv
     \bigvee_{j,j'}(\pvec{y}= \pvec{r}^{sa_i}_j(\pvec{x}\,) \land 
     \phi^{sa_i}_{j'}(\pvec{x}\,)
     \land v=c^{sa_i}_{j,j'}(\pvec{x}\,) )  \\
    &\mathcal{L}_{sen_i}(\pvec{y}\,)=v \equiv 
     \bigvee_{j,j'}
     (\pvec{y}= \pvec{r}^{sen_i}_j \land 
     \phi^{sen_i}_{j'} \land v=c^{sen_i}_{j,j'} )
     \end{aligned}
\)
}
$$


\noindent here  $\pvec{r}^{sa_i}_j(\pvec{x}\,)$ and $c^{sa_i}_{j,j'}(\pvec{x}\,)$  are rigid terms with variables $\pvec{x}$;
$\phi^{sa_i}_{j'}(\pvec{x}\,)$, the \emph{likelihood contexts}, are fluent formulas with free variables among $\pvec{x}$;  
$\pvec{r}^{sa_i}_j$ and $c^{sen_i}_{j,j'}$ are rigid terms, $\phi^{sa_i}_{j'} $ are fluent formula without variables. Besides, we require that likelihood contexts are disjoint and complete: 1) for all \(i\) and distinct \(j'_1,j'_2\)
    \(\models \all \pvec{x}. (\phi^{sa_i}_{j'_1}(\pvec{x}\,) \supset \neg \phi^{sa_i}_{j'_2}(\pvec{x}\,))\);
   2) \( \models \all \pvec{x}. \bigvee_{j'} \phi^{sa_i}_{j'}(\pvec{x}\,)\) for all $i$; 
    3) \( \models \sum_{j} c^{sa_i}_{j,j'} = 1\)  for all \(i,j'\).
     
\item $\bel^f$ in $\KB$ is finite, namely, of the form $f(\pvec{u})=v \equiv \bigvee_i \pvec{u} = \pvec{n}_i \land v=r_i $ and $\sum_{i} r_i =1$.    
 \end{enumerate}

 Intuitively, the first two conditions ensure that for any $sa_i(\pvec{x},\pvec{y}\,)$, only finitely many alternatives, which satisfy $\pvec{y} = \pvec{r}^{sa_i}_j(\pvec{x}\,)$, have non-zero likelihood; similarly, sensing only has finitely many outcomes: $\pvec{y} = \pvec{r}^{sen_i}_j $.
 The third item says that only finitely many fluent values are believed with non-zero degree. With these restrictions, $\sum_{\pvec{u}}f(\pvec{u})$ can be replaced by the finite sum $\sum_{i}f(\pvec{n}_i)$ and $\sum_{a \in \N_A} \mathcal{L}(a)$ can be replaced by the finite sum $\sum_{j}\sum_{j'} c^{sen_i}_{j,j'}$. The BAT and $\KB$ in Example~\ref{example:bat} satisfy all the above conditions. A remark is that given a $\KB$ with a finite belief distribution and a BAT satisfying the above conditions, the belief distribution of its progression is still finite.

\section{The Proposed Framework}
\label{sec:framework}

\subsection{Belief Programs}

The atomic instructions of our belief programs are the so-called \emph{primitive programs} which are actions that suppress their uncontrollable parameters. A primitive program $\pp$ can be \emph{instantiated} by a ground action $t_a$, i.e. $\pp \rightarrow t_a$, iff $\Sigma_{oi} \models \exists \pvec{y}.oi(\pp[\pvec{y}],t_a) = \top$, where \(\pp[\pvec{y}]\) is the action that restores its suppressed parameters by \(\pvec{y}\). For instance, \(east(1) \rightarrow east(1,1)\), \(sencfe \rightarrow sencfe(1)\).


\begin{definition}
A \textbf{program expression} \(\delta\) is defined as :
\[
\delta ::= \pp| \alpha?| (\delta;\delta) | (\delta|\delta)|\delta^*
\]
\end{definition}

Namely, a program expression can be a primitive program \(\pp\), a test \(\alpha?\) where \(\alpha\) is a static subjective formula without $\oknow$, or constructed from sub-program by sequence \(\delta;\delta\), non-deterministic choice \(\delta|\delta\), and non-deterministic iteration \(\delta^*\). Furthermore, \textbf{if} statements and \textbf{while} loops can be defined as abbreviations in terms of these constructs:
$$
\begin{aligned}
& \textbf{if}~\alpha~\textbf{then}~\delta_1~\textbf{else}~\delta_2~\textbf{endIf} := [\alpha?;\delta_1]|[\neg \alpha?;\delta_2]\\
& \textbf{while}~\alpha~\textbf{do}~\delta~\textbf{endWhile} := [\alpha?;\delta]^*;\neg \alpha?
\end{aligned}
$$

Given BATs $\Sigma,\Sigma'$, the initial state axioms \(\Sigma_0\), a $\KB$ $\oknow(\bel^f \land \Sigma')$, and a program expression \(\pe\), a \textbf{belief program} \(\calP\) is a pair \(\calP = (\Sigma_0 \union \Sigma \union \oknow(\bel^f\land \Sigma'), \pe )\). An example of a belief program is $\calP$ where $\pe$ is given by Table.~\ref{tab:bpexample} and $\Sigma_0$, $\Sigma$, $\KB$ are given by Example.~\ref{example:bat}. \footnote{
We use \(\bel(h=2) < 1\) to denote \(\exists u.\bel(h=2 \colon u) \land u< 1\). 
The confidence $\conf(h,u)$ of a fluent $h$ of sort number wrt $u$ is defined as: $\Box \conf(h,u) = v \equiv \bel(|h-\exp(h)|<u \colon v)$ while the expectation $\exp(h)$ is defined as $\Box \exp(h) = v \equiv v=\sum_{u \in \cn} u \times~(\textbf{if}~\exists u'. \bel(h=u \colon u')~\textbf{then}~v~\textbf{else}~0)
$.}

In order to handle termination and failure, we reserve two nullary fluents \(Final\) and \(Fail\). Moreover, \(\Box[a]Final = u \equiv a = \e \land u=\top \vee Final = u\) (likewise for \(Fail\) with action \(\f\)) is implicitly assumed to be part of \(\Sigma\) and \(\Sigma'\). Additionally, \(\Sigma_0 \models Final = \bot \land Fail=\bot\),
and actions $\e,\f$ do not occur in $\pe$. 
A \emph{configuration} \(\lan z,\delta \ran \) consists of an action sequence \(z\) and a program expression \(\delta\).

\begin{definition}[program semantics] 
Let \(\calP = (\Sigma_0 \union \Sigma \union \oknow(\bel^f \land \Sigma'), \pe )\) be a belief program, the transition relation \(\trans\) among configurations, given $e$ s.t. $e \models \oknow(\bel^f \land \Sigma')$, is defined inductively:

\begin{enumerate}
    \item \(\lan z,\pp \ran \trans \lan z\cdot t, \lan \ran \ran \), if  \(\pp \rightarrow t\);
    
    \item \(\lan z,\delta_1;\delta_2 \ran \trans \lan z\cdot t, \delta'; \delta_2 \ran\), if \(\lan z,\delta_1 \ran \trans \lan z \cdot t, \delta' \ran \);
    
    \item \(\lan z,\delta_1;\delta_2 \ran \trans \lan z\cdot t, \delta' \ran\), if \(\lan z,\delta_1 \ran \in \Fin\) and \(\lan z,\delta_2 \ran  \trans \lan z \cdot t, \delta' \ran \);
    
    \item \(\lan z,\delta_1|\delta_2 \ran \trans \lan z\cdot t, \delta' \ran\), if \(\lan z,\delta_1 \ran \trans \lan z\cdot t, \delta' \ran\) or \(\lan z,\delta_2 \ran \trans \lan z\cdot t, \delta' \ran\) ;
    
    \item \(\lan z,\delta^* \ran \trans \lan z\cdot t, \delta';\delta^* \ran\), if \(\lan z,\delta \ran \trans \lan z\cdot t,\delta' \ran \) .
\end{enumerate}


The set of final configuration \(\Fin\) wrt \(e\) is the smallest set such that:

\begin{enumerate}
    \item \( \lan z,\lan \ran \ran \in \Fin\);
    
    \item \( \lan z,\alpha? \ran \in \Fin\) if \(e,w, z \models \alpha\);
    
    \item \( \lan z, \delta_1;\delta_2 \ran \in \Fin\) if  \( \lan z, \delta_1\ran \in \Fin\) and \( \lan z, \delta_2\ran \in \Fin\);
    
    \item \( \lan z, \delta_1|\delta_2 \ran \in \Fin\) if  \( \lan z, \delta_1\ran \in \Fin\) or \( \lan z, \delta_2\ran \in \Fin\);
    
    \item \( \lan z, \delta^* \ran \in \Fin\);
    
\end{enumerate}



\end{definition}
The set of \emph{failing configurations} is given by:
$\Fail = \{ \lan z,\delta \ran |  \lan z,\delta \ran \notin \Fin,~\textrm{there is no }~\lan z\cdot t,\delta' \ran ~\textrm{s.t.}~ \lan z,\delta \ran \trans \lan z\cdot t,\delta' \ran  \}$. 

 We extend final and failing configurations with addition transitions. This is achieved by defining an extension of $\trans$. The \emph{extended transition relation} $\etrans$ \emph{among configurations} is defined as the least set such that:

\begin{enumerate}
    \item \(\lan z,\delta\ran \etrans \lan z \cdot t, \delta' \ran \) if \(\lan z,\delta\ran \trans \lan z \cdot t, \delta' \ran \);
    
    \item \(\lan z,\delta\ran \etrans \lan z \cdot \e, \lan \ran \ran \) if \(\lan z,\delta\ran \in \Fin\);
    
  \item \(\lan z,\delta\ran \etrans \lan z \cdot \f, \lan \ran \ran \) if \(\lan z,\delta\ran \in \Fail\).
\end{enumerate}

The execution of a program \(\calP\) yields a countably infinite~\footnote{Our restrictions on \(\Sigma_{oi}\) and \(\Sigma_{l}\) ensure a bounded branching for the MDP, therefore its states are countable.} \emph{Markov Decision Process} \(\mathsf{M}^{e,w}_{\pe}=(\mathsf{S},\mathsf{A},\mathsf{P},s_0)\)  wrt \(e,w\) s.t. \(e,w \models \Sigma_0 \union \Sigma \union \oknow(\bel^f \land \Sigma')\).

\begin{enumerate}

\item \(\mathsf{S}\) is the set of configurations reachable from \(\lan \lan \ran, \pe \ran \) under  \(\etrans^*\) (transitive and reflexive closure of  \(\etrans\));

\item \(\mathsf{A}\) is the finite set of primitive programs in \(\pe\);

\item \(\mathsf{P}\) is the transition function \(\mathsf{P} \colon \mathsf{S} \times \mathsf{A}\times \mathsf{S} \rightarrow \cn \)

with \(\mathsf{P}(\lan z, \delta \ran, \pp, \lan z\cdot t, \delta' \ran)\) given by:

\(
\mathsf{P}(\cdot) =\left\{
\begin{array}{cc}
     p &  \begin{array}{c}
         \pp \rightarrow t~\textrm{,}~w,z \models l(t) = p,   \\
            ~\textrm{and}~ \lan z, \delta \ran \etrans  \lan z\cdot t, \delta' \ran
     \end{array}
     \\
    1  & \lan z,\delta \ran \in \Fin
         \textrm{~and~} \pp = t=\delta'=\e
    \\
    1  & \lan z,\delta \ran \in \Fail \textrm{~and~} \pp = t=\f, \delta'=\delta
    \\
    0 & o.w.
\end{array}
\right.
\)


\item \(s_0\) is the initial state \(\lan \lan \ran, \pe \ran \).

\end{enumerate}

Now, the non-determinism on the agent's sides is resolved by means of \emph{policy} \(\sigma\), which is a mapping \(\sigma: \mathsf{S}  \mapsto \mathsf{A}\). 
A policy \(\sigma\) is said to be \emph{proper} if and only if for all \(s=\lan z, \delta \ran\), \(s'=\lan z', \delta' \ran\), 
if $\models Pro(\oknow(\bel^f\land \Sigma'),z) \equiv Pro(\oknow(\bel^f\land \Sigma'),z')$ then \(\sigma(s) = \sigma(s')\), namely, the robot acts only according to its $\KB$.
An infinite path \(\pi= s_0 \xrightarrow{\pp_1} s_1 \xrightarrow{\pp_2} s_2\cdots \) is called a \(\sigma\textrm{-}path\) if \(\sigma(s_j)=\pp_j\) for all \(j\ge 0\). The \(j\)-th state of any such path is denoted by \(\pi[j]\). The set of all \(\sigma\textrm{-}paths\)  starting in \(s\) is denoted by \(\mathsf{Path}^\sigma(s,\MDP^{e,w}_{\pe})\).

Every policy \(\sigma\) induces a probability space \(\mathsf{Pr}^\sigma_s\) on the set of infinite paths starting in \(s\), using the cylinder set construction: For any finite path prefix \(\pi_{\mathsf{fin}} =  s_0 \xrightarrow{\pp_1} s_1 \cdots s_n \), we define the probability measure:
$$\mathsf{Pr}^{\sigma}_{s_0,\mathsf{fin}} = \mathsf{P}(s_0,\pp_1,s_1) \times \mathsf{P}(s_1,\pp_2,s_2) \cdots \mathsf{P}(s_{n-1},\pp_{n},s_n) $$

\subsection{Temporal Properties of Programs}
We use a variant of PCTL to specify program properties. The syntax is given as:
\begin{gather}
\Phi ::= \beta | \neg \Phi | \Phi \land \Phi | \mathbf{P}_{I}[\Psi]   \tag{A} \\
\Psi ::= \mathbf{X}\Phi | (\Phi\mathbf{U}\Phi) | (\Phi\mathbf{U}^{\le k}\Phi)  \tag{B}
\end{gather}
\noindent where \(\beta\) is a static subjective $\lang$ formula without $\oknow$. 
We call formulas according to (A) state formulas and according to (B) trace formulas. Here \(I \subseteq [0,1]\) is an interval. \(\Phi\mathbf{U}^{\le k}\Phi\)
is the step-bounded version of the until operator. Some useful abbreviations are:
\(\mathbf{F}\Phi\)
(eventually \(\Phi\)) for \(\true \mathbf{U} \Phi\)
and \(\mathbf{G}\Phi\) (globally \(\Phi\))
for \(\neg \mathbf{F} \neg \Phi\).


	    
	
	
Let \(\Phi\) be a temporal state formula, \(\Psi\) a temporal trace formula,  \(\MDP^{e,w}_{\pe}\) the infinite-state MDP of a program  \(\calP = (\Sigma_0 \union \Sigma \union \oknow(\bel^f \land \Sigma'), \pe )\) wrt \(e,w\) s.t. \(e,w \models \Sigma_0 \union \Sigma \union \oknow(\bel^f \land \Sigma')\), and \(s \in \mathsf{S} \).
Truth of state formula $\Phi$ is given as:
	\begin{enumerate}
	    \item \(\MDP^{e,w}_{\pe},s \models \beta \)  iff \(s=\lan z,\delta\ran\) and \(e,w,z \models \beta\) ;
	    
	   \item \(\MDP^{e,w}_{\pe},s \models \neg \Phi \)  iff \(\MDP^{e,w}_{\pe},s \nvDash  \Phi\) ;
	   
	   \item \(\MDP^{e,w}_{\pe},s \models \Phi_1 \land \Phi_2 \)  iff \(\MDP^{e,w}_{\pe},s \models \Phi_1\) and \(\MDP^{e,w}_{\pe},s \models \Phi_2\) ;
	    
	    \item \(\MDP^{e,w}_{\pe},s \models \mathbf{P}_{I}[\Psi]\)  iff for all \emph{proper} policies \(\sigma\), \(\mathsf{Pr}^{\sigma}_{s}(\Psi) \in I\), where
\[ \mathsf{Pr}^{\sigma}_{s}(\Psi)= \mathsf{Pr}^{\sigma}_{s}(\{\pi\in \mathsf{Path}^\sigma(s,\MDP^{e,w}_{\delta}) |\MDP^{e,w}_{\delta},\pi \models \Psi \}) .\]
	\end{enumerate}  

Furthermore, let $\pi \in \mathsf{Path}^\sigma(s,\MDP^{e,w}_{\pe})$ be an infinite path for some proper policy $\sigma$, truth of trace formula $\Psi$ is as:

\begin{enumerate}
    \item \(\MDP^{e,w}_{\pe},\pi\models \mathbf{X}\Phi\) iff  \(\MDP^{e,w}_{\pe},\pi[1] \models \Phi\);
     
     \item \(\MDP^{e,w}_{\pe},\pi\models \Phi_1 \mathbf{U} \Phi_2\) iff  \(\exists i. 0 \le i \) s.t. \( \MDP^{e,w}_{\pe},\pi[i] \models \Phi_2\) and \(\all j. 0\le j \le i,     \MDP^{e,w}_{\pe},\pi[j] \models \Phi_1\);
     
      \item \(\MDP^{e,w}_{\pe},\pi\models \Phi_1 \mathbf{U}^{\le k} \Phi_2\) iff  \(\exists i. 0 \le i \le k\) s.t. \( \MDP^{e,w}_{\pe},\pi[i] \models \Phi_2\) and \(\all j. 0\le j \le i,     \MDP^{e,w}_{\pe},\pi[j] \models \Phi_1\);
    
\end{enumerate}

\begin{definition}[Verification Problem] 
\label{def:verification}
A temporal state formula \(\Phi\) is valid in a program \(\calP\), \(\calP \models \Phi\), iff  for all \(e,w\) with \(e,w \models \Sigma_0 \union \Sigma \union \oknow(\bel^f \land \Sigma')\), it holds that  \(\MDP^{e,w}_{\pe},s_0 \models \Phi \).
\end{definition}

E.g. $\mathbf{P}_{\ge 0.05}[\mathbf{F}^{\le 2} \bel(h=2 \colon 1)]$ and $\mathbf{P}_{=1}[\mathbf{F} \bel(h=2 \colon 1)]$ specify the two properties \textbf{P1} and \textbf{P2} in the introduction respectively.

\section{Undecidability}
\label{sec:undecidability}

The verification problem is undecidable because belief programs are probabilistic variants of $\golog$ programs with sensing, for which  undecidability was shown in ~\cite{zarriess2016decidable}. 
Cla{\ss}en et al.~\shortcite{classen2013decidable} observed that many dimensions affect the complexity of the $\golog$ program verification including the underlying logic, the program constructs, and the domain specifications. Since then, efforts have been made to find decidable fragments. Arguably, the dimension of domain specification is less well-studied. Here we study the boundary of decidability from this dimension. Hence, in this paper, we set the other two dimensions to a known decidable status.\footnote{
 Formally, we assume our logic only contains $+,\times$ as rigid function symbols and whenever we write logical entailment $\Sigma \models \alpha$, we mean $\Sigma \union \Delta \union \T_{\R}\models \alpha$ where  $\Delta$ is as before and  $\T_{\R}$ is the theory of the reals,
 where validity is decidable~\cite{tarski1998decision}. In terms of program constructs, we disallow non-deterministic pick of program parameters, $\pi x. \delta (x) $, which is proven to be a source of undecidability in~\cite{classen2013decidable}.
}

In deterministic settings, domain specifications mainly refer to SSAs. Nevertheless, in our case, the likelihood axiom (LA) plays an important role as well. Some relevant variants of SSAs are \emph{context-free} \cite{reiter:knowledge} and \emph{local-effect} SSAs \cite{liu2005tractable}.

\begin{definition}
 A set of SSAs is called:
 \begin{enumerate}
     \item context-free, if for all fluents $h$, \(\gamma_h\) is rigid;
     \item local-effect, if for all fluents $h$, \(\gamma_h \) is a  disjunction of the form \(\exists \vec{\mu}. a = as(\pvec{v}) \land \nabla \), where
     \(as\) is an action symbol,  \(\pvec{v}\) contains \(u\) and \(\mu\), and \(\nabla\) is a fluent formula with free variables in \(\pvec{v}\).
 \end{enumerate}
\end{definition}

Intuitively, context-free means that effects of actions are independent of the state while for local-effect, effects might depend on the state specified by \emph{effect context} $\nabla$  but only locally. An example of local-effect SSAs is the blocks-world domain, where the action $move(x, y, z)$, i.e.
moving object $x$ from $y$ to $z$, only affects properties of objects $x,y,z$.  The SSA in Example~\ref{example:bat} is not local-effect. A \(\textrm{context-free SSA}\) is also \( \textrm{local-effect}\).

Since $\gamma_{h}$ is functional in $u$ and only finitely many action symbols are used: $sa_1,\ldots, sa_k$ for stochastic actions or $sen_1,\ldots, sen_{k'}$ for sensing, $\gamma_{h}$ can be written in the form (sensing does not change fluents):
\begin{equation}
 	     \label{eq:ssaRstr}
	\begin{aligned}\gamma_{h} \equiv& \bigvee_i \exists \pvec{x},\pvec{y}. a = sa_i(\pvec{x},\pvec{y}\,) \land u = t^{sa_i}_{h}(\pvec{x},\pvec{y}\,)\\
	& \vee \all \pvec{x},\pvec{y}. \band_i a \neq sa_i(\pvec{x},\pvec{y}\,) \land h=u
 	\end{aligned}
 	 \end{equation}
	where $t^{sa_i}_{h}(\pvec{x},\pvec{y}\,)$ are definitional terms with variables \(\pvec{x},\pvec{y}\). After such rewrite, a SSA is context-free iff $t^{sa_i}_{h}(\pvec{x},\pvec{y}\,)$ are rigid. To ensure a SSA to be local-effect, we require that the $t^{sa_i}_{h}(\pvec{x},\pvec{y}\,)$ in Eq.~(\ref{eq:ssaRstr}) are of the form:
$$t^{sa_i}_{h}(\pvec{x},\pvec{y}\,) = \left\{ \begin{array}{cc}
    v_1 &  \nabla_1(\pvec{x},\pvec{y}\,)\\
    \vdots\\
    v_k & \nabla_k(\pvec{x},\pvec{y}\,)
\end{array}\right.$$

\noindent where $v_i$ are variables among $\pvec{x}\union \pvec{y}$ and $\nabla_i(\pvec{x},\pvec{y}\,)$, the effect contexts, are fluent formulas with free variables among $\pvec{x}\union\pvec{y}$. Obviously, this restriction is sufficient to ensure the SSA to be local-effect: since $u=v_i$ for some $v_i \in \pvec{x} \union \pvec{y}$, the variable $v_i$ can be eliminated by replacing it with $u$ directly, which further ensures the SSA fulfills the definition of local-effect. 

We call a LA \emph{context-free} if the RHS of \(\Sigma_{l}\) is rigid. Obviously, context-free LA excludes sensing since sensing always involves fluents.







Table~\ref{tab:undecidability} lists the decidability of the belief program verification problem. Dashes mean no constraint. The result is arranged as follows. We first explore decidability for the case with no restriction on the LA. As it turns out, the problem is undecidable even if SSAs are context-free (1). Therefore, we set the LA to be context-free, which results in undecidability for the case of local-effect SSAs (2). The case with question mark remains open (3). 

\begin{table}[t]
    \centering
    \begin{tabular}{|c|c|c|c|}
    \hline
     \# & LA & SSA & Decidable \\
     \hline
    1 & -  &  context-free & No \\
    2 & context-free  & local-effect &  No\\
    3 & context-free &  context-free & ?\\
     \hline
    \end{tabular}
    \caption{Decidability of the verification problem}
    \label{tab:undecidability}
\end{table}

\begin{theorem}
\label{theorem:undcf}
The verification problem is undecidable for programs with context-free SSAs.
\end{theorem}


\begin{hproof}
We show the undecidability by a reduction of the undecidable \emph{emptiness problem of probabilistic automata} ~\cite{paz2014introduction}.
A \emph{probabilistic finite automaton} (PA) is a quintuple \(\A = ( \mathsf{Q}, \mathsf{L},\left(\mathsf{M}_{l}\right)_{l \in \mathsf{L}}, q_{1}, \mathsf{F} )\)  where \(\mathsf{Q}\) is a finite set of states,
\(\mathsf{L}\) is a finite alphabet of letters, \((\mathsf{M}_{l})_{l \in \mathsf{L}}\) are the stochastic transition matrices, \(q_{1}\in \mathsf{Q}\) is the initial state and \(\mathsf{F} \subseteq \mathsf{Q}\) is a set of accepting states.
For each letter \(l \in \mathsf{L}, \mathsf{M}_{l} \in[0,1]^{\mathsf{Q} \times \mathsf{Q}}\) defines transition probabilities: \(0 \leq \mathsf{M}_{l}(q_i,q_j) \leq 1\) is the probability from
state \(q_i\) to \(q_j\) when reading a letter \(l\).
The \emph{emptiness problem} is that given a PA $\A$ and $\xi \in [0,1]$, deciding whether there exists a word $\wo $ (a sequence of letters) such that
$\mathbb{P}_{\mathcal{A}}(q_{1} \stackrel{\wo}{\rightarrow} \mathsf{F} ) \ge \xi$, namely, the probability of reaching accepting states from the initial state upon reading $\wo$ is no less than $\xi$. The emptiness problem is known to be undecidable. The following is a belief program with context-free SSAs to simulate the run of a given probabilistic finite automaton $\A$ and threshold $\xi$.

Formally, we have a single fluent $h_s$ to record the current state, a set of standard names \( \N_{\mathsf{Q}}=\{n_1,n_2\ldots, n_{|\mathsf{Q}|}\}\) to represent the states in $\mathsf{Q}$, a set of stochastic actions \(\pp_{i}(y)\) to simulate the read of letter \(l_i \in \mathsf{L}\). For the BAT $\Sigma$, we have 

\(
    \begin{aligned}
     &\Box[a] h_s=u \equiv \bigvee_{i} \exists y. a=\varrho_{i}(y) \land u=y \\
     &\quad\quad\quad\quad\quad\vee \band_i \all y. a \neq \varrho_{i}(y) \land h_s=u \\
     &\Box l(a) = v \equiv \bigvee_{i} \exists y. a=\varrho_{i}(y) \land v=\mathcal{L}_{\varrho_i}(y) 
     \end{aligned} 
\)

\noindent where $\mathcal{L}_{\varrho_i}(y)$ is given by
\begin{equation}
    \label{eq:LaReduction1}
    \begin{aligned}
    \mathcal{L}_{\varrho_i}(y)  = \left \{ 
    \begin{array}{cc}
        \mathsf{M}_{l_i}(h_s,y) & h_s,y \in \N_{\mathsf{Q}} \\
        0 & o.w.
    \end{array}
    \right.
    \end{aligned}
\end{equation}

Intuitively, the BAT says that fluent $h_s$ can only be changed by action $\varrho_{i}(y)$ and the unobservable parameter $y$ determines the new state; the likelihood of $\varrho_{i}(y)$ depends on the current state $h_s$ and equals the transition probability $\mathsf{M}_{l_i}(h_s,y)$.
Now let $\Sigma_0 = \{h_s=n_1\}$, $\bel^f\equiv\bel(h_s=n_1 \colon 1)$, then the program $\calP=(\Sigma_0\union\Sigma \union \oknow(\bel^f \land \Sigma), \pe)$ simulates the run of PA $\A$ where
\[\pe::= \textbf{while}~ \bel(h_s \in \N_{\mathsf{F}})<\xi~\textbf{do}~\pp_{1}\left|\pp_{2}, \ldots,\right| \pp_{|\mathsf{L}|} ~\textbf{endwhile}.\]

\noindent Here $\N_{\mathsf{F}} $ is the set of standard names representing the accepting states $\mathsf{F}$ in $\A$ and $\pp_{i}\rightarrow \varrho_{i}(y)$. This is sound in the sense that for any action sequence $z$ composed by ground actions in $\pp_i(y)$ s.t. $y \in \N_{\mathsf{Q}}$, 
and any number $r$, $\oknow(\bel^f \land \Sigma) \models[z]\bel(h_s \in \N_{\mathsf{F}} \colon r)$ iff $\mathbb{P}_{\mathcal{A}}(q_{1} \stackrel{\wo}{\rightarrow} \mathsf{F} ) = r$, where $\wo$ is the corresponding word of $z$. Hence, 

\noindent
\resizebox{1\linewidth}{!}{
$\calP \models \mathbf{P}_{>0}[\mathbf{F} (\bel(h_s \in \N_{\mathsf{F}}) \ge \xi)] \textrm{ iff } \exists \wo.\mathbb{P}_{\mathcal{A}}(q_{1} \stackrel{\wo}{\rightarrow} \mathsf{F} ) \ge \xi$
}
\end{hproof}

A crucial point in the above reduction is that the RHS of the likelihood axiom is not rigid, which further allows us to specify action likelihood according to transition probabilities of a PA ($h_s$ in Eq.~(\ref{eq:LaReduction1})). A natural question is whether the verification problem is decidable if we set the LA to be rigid. The following theorem provides a negative answer for this when the SSAs are local-effect.

\begin{theorem}
The verification problem is undecidable for programs with local-effect SSAs and context-free LA.
\end{theorem}

Since the LA is restricted to be context-free, the previous reduction breaks as transition probabilities of probabilistic automata might depend on states in general. Nevertheless, we reduce the emptiness problem of the \emph{simple probabilistic automata} (SPA), i.e. PA whose transition probabilities are among $\{0,\frac{1}{2},1\}$, to the verification problem with context-free LA and local-effect SSAs. More precisely, the simple probabilistic automata we considered are \emph{super simple probabilistic automata} (SSPA), SPA with a single probabilistic transition and every transition has a unique letter. Fijalkow et al.~\shortcite{fijalkow2012deciding} show that the emptiness problem of the SPA with even a single probabilistic transition is undecidable. Their result can be easily extended to SSPA.

The idea of the reduction is to shift the likelihood context in LAs to the context formula in SSAs. More concretely, instead of saying an action's likelihood depends on the state and the action's effect is fixed, which is the view of the BAT in the previous reduction, we say the action's effect depends on the state and the action's likelihood is fixed. This is better illustrated by an example. Consider a SSPA consisting of a single probabilistic transition
with $q \xrightarrow{0.5,l} q'$ and $q \xrightarrow{0.5,l} q''$, clearly, one can construct a BAT as in the previous reduction to simulate this, nevertheless, 
the following BAT with a local-effect SSA and a context-free LA can simulate it as well:

 \(
    \begin{aligned}
     & \Box[a] h_s=u \equiv \exists y. a=\varrho(y) \land u=y \land h_s=n\\
       &\quad\quad\quad\quad\quad\quad\vee ( \all y. a \neq \varrho(y) \vee h_s \neq n)\land h_s=u \\
     &\Box l(a) = v \equiv  \exists y. a=\varrho(y) \land v= \left\{ 
     \begin{array}{cc}
         \frac{1}{2} &  y \in \{n,n''\}\\
         0 & o.w.
     \end{array}
     \right. 
    \end{aligned} 
\)

\noindent Here, $n,n'n''$ are standard names corresponding to the states $q,q',q''$. The SSA is local-effect as it complies with our conditions for local-effect SSAs: $t^{\pp}_{h_s}(y) = y$ and $\nabla(y) \equiv h_s=n$.
The simulation is sound in the sense that the 
belief distribution of fluent $h_s$ corresponds to the probability distribution among states, as in the previous reduction.


\section{A Decidable Case}
\label{sec:decidability}
Another source of undecidability comes from the property specification, more precisely, the unbounded until operators. In fact, in our program semantics, the MDP $\mathsf{M}^{e,w}_{\pe}$ is indeed an infinite partially observable MDP (POMDP) where the set of observations is just the set of possible $\KB$s that can be progressed to from the initial $\KB$ regarding a certain possible action sequence of the program. Verifying belief programs against specifications with unbounded $\mathbf{U}$ requires verification of indefinite-horizon POMDPs, which is known to be undecidable. This motivates us to focus on the case with only bounded until operators. In contrast to the previous section, we now allow arbitrary domain specifications.

A state formula $\Phi'$ is called \emph{bounded} iff it contains no $\mathbf{U}$ and no nested $\mathbf{P}$, namely, $\Phi' ::= \beta | \mathbf{P}_{I}[\Psi']$ with $\Psi' ::= \mathbf{X}\beta | (\beta\mathbf{U}^{\le k}\beta) $.\footnote{Verifying properties with nested $\mathbf{P}$ is known to be considerably more difficult \cite{norman2017verification}.}

For example, the property \textbf{P1}
$\mathbf{P}_{\ge 0.05}[\mathbf{F}^{\le 2} \bel(h=2 \colon 1)]$ is bounded while the property \textbf{P2} is not. For bounded state formulas, we only need to consider action sequences with a bounded length, namely, only a finite subset of $\mathsf{M}^{e,w}_{\pe} $'s states and observations needs to be considered. Although model-checking the finite subset of $\mathsf{M}^{e,w}_{\pe} $ against PCLT formulas without unbounded $\mathbf{U}$ operators is decidable, this does not entail that the verification problem is decidable as infinitely many such subsets exist. This is because there are infinitely many models $(e,w)$ satisfying the initial state axioms. Our solution is to abstract them into finitely many equivalence classes \cite{zarriess2016decidable}.

First, we need to identify the so-called \emph{program context} $\C(\calP)$ of a given program $\calP$, which contains: 1) all sentences in $\Sigma_0$; 2) all likelihood conditions $\phi^{sa_i}_{j'}(\pvec{x}\,)$ and $\phi^{sen_i}_{j'}(\pvec{x}\,)$; 3) all test conditions in the program expression; 4) all $\lang$ sub-formulas in the temporal property; 5) the negation of formulas from 1) - 4).
We then define types of models as follows:
\begin{definition}[Types] Given a belief program $\calP$ and a bounded state formula $\Phi'$, let $\A_{\calP}$ be the set of all ground actions with non-zero likelihood in $\calP$, $(\A_{\calP})^k$ be the set of all action sequences by actions in $\A_{\calP}$ with length no greater than $k$.\footnote{
If $\Phi'$ does not contain bounded until operators, we set $k=0$ for $\Phi'\equiv \beta$ and $k=1$ for $\Phi'\equiv\mathbf{P}_{I}[\mathbf{X}\beta] $.} The set of all type elements is given by:

$\mathsf{TE}(\calP,\Phi') = \{(z,\alpha) | z\in (\A_{\calP})^k, \alpha \in \C(\calP)\}$

\noindent A type wrt $\calP,\Phi'$ is a set $\tau \subseteq \mathsf{TE}(\calP,\Phi')$ that satisfies:

\begin{enumerate}
    \item $\all \alpha \in \C(\calP)$, $ \all z\in (\A_{\calP})^k$, $(z,\alpha) \in \tau$ or  $(z,\neg \alpha) \in \tau$;
    
    \item there exists $e,w$ s.t. $e,w \models \Sigma_0 \union \oknow(\bel^f \land \Sigma') \union \{[z]\alpha~|~ (z,\alpha) \in \tau \}$. 
\end{enumerate}
\end{definition}

Let $\mathsf{Types}(\calP,\Phi')$ denote the set of all types wrt $\calP$ and $\Phi'$. The type of a model $(e,w)$ is given by
$
\mathsf{type}(e,w) ::=\{(z,\alpha)\in \mathsf{TE}(\calP,\Phi')~|~ e,w \models [z]\alpha\}
$. $\mathsf{Types}(\calP,\Phi')$ partitions $e,w$ into equivalence classes in the sense that if $\mathsf{type}(e,w) = \mathsf{type}(e',w')$, then $e,w \models [z]\alpha$ iff 
$e',w' \models [z]\alpha$ for $z\in (\A_{\calP})^k $ and $\alpha \in \C(\calP)$.


Thirdly, we use a representation similar to the \emph{characteristic program graph} \cite{classen2008logic} where nodes are the reachable subprograms $Sub(\pe)$, each of which is associated with a termination condition $\mathsf{Fin}(\delta')$ (the initial node $v_0$ corresponds to the overall program $\pe$), and where an edge $\delta_1 \xrightarrow{\pp/\alpha} \delta_2$ represents a transition from $\delta_1$ to $\delta_2$
by the primitive program $\pp$ if test condition $\alpha$ holds. Moreover, failure conditions are given by 
$\mathsf{Fail}(\delta') ::= \neg( \mathsf{Fin}(\delta') \vee \bigvee_{\delta' \xrightarrow{\pp/\alpha} \delta''} \alpha )$. 

Lastly, we define a set of atomic propositions $\mathsf{AP}= \{p_{\alpha} | \alpha \in \C(\calP)\textrm{~and~} \alpha\textrm{~is subjective}\} $ one for each subjective $\alpha \in \C(\calP)$.

The finite POMDP for a type $\tau$ of a program $\calP$ is a tuple $\mathsf{M}^{\tau}_{\pe} = \lan \mathsf{S_{fin}},\mathsf{s^0_{fin}},\mathsf{A_{fin}}, \mathsf{P_{fin}},\mathsf{O_{fin}},\mathsf{\Omega_{fin}}, \mathsf{L_{fin}} \ran$ consisting of: 

\begin{enumerate}
 \setlength\itemsep{-0.2em}
    \item the set of states $\mathsf{S_{fin}} = (\A_{\calP})^k \times Sub(\pe)$;
    
    \item the initial state $\mathsf{s^0_{fin}} = \lan \lan \ran,\pe \ran$;
    
    \item the set of primitive programs $\mathsf{A_{fin}} = \mathsf{A}$;
    
    \item the transition function $\mathsf{P_{fin}}(\lan z_1, \delta_1 \ran, \pp, \lan z_2 , \delta_2 \ran )$ as 
    \begin{itemize}
        \item $\mathsf{P_{fin}}(\cdot) = c^{sa_i}_{j,j'}(\pvec{n}) $ if 
    $|z_1| < k$, $\delta_1 \xrightarrow{\pp/\alpha} \delta_2$, $(z_1,\alpha) \in \tau$, and for some $sa_i,\pvec{n}$, 
    $ \pvec{r}^{sa_i}_j$, $\phi^{sa_i}_{j'}(\pvec{n})$, it holds that (likewise for sensing)
   $$
   \resizebox{1\linewidth}{!}{
$\pp \rightarrow sa_i(\pvec{n},\pvec{r}^{sa_i}_j),z_2 = z_1 \cdot sa_i(\pvec{n},\pvec{r}^{sa_i}_j),  (z_1,\phi^{sa_i}_{j'}(\pvec{n})) \in \tau;$
}$$

    \item $\mathsf{P_{fin}}(\cdot) = 1 $ if $|z_1| = k,\pp=\f, z_1 = z_2, \delta_2 = \delta_1$;
    
    \item $\mathsf{P_{fin}}(\cdot) = 1 $ if  $(z_1, \mathsf{Fin}(\delta_1)) \in \tau, \pp=\delta_2=\e$ ;
    
    \item $\mathsf{P_{fin}}(\cdot) = 1 $ if  $(z_1, \mathsf{Fail}(\delta_1)) \in \tau, \pp=\delta_2=\f$ ;
    
    \end{itemize}
    
    \item  the observations $\mathsf{O_{fin}} = \{Pro(\oknow(\bel^f\land \Sigma'),z) | z \in \A_{\calP}^k \}$;
    
    \item  the state to observation mapping $\mathsf{\Omega_{fin}}$ as $\mathsf{\Omega_{fin}}(\lan z,\delta \ran) = Pro(\oknow(\bel^f\land \Sigma'),z)$;
    
    \item the labeling $\mathsf{L_{fin}}(o) = \{ p_{\alpha} | p_{\alpha} \in \mathsf{AP}, o \models \alpha \}$. \footnote{Here, we use a function $\textbf{E}[\KB,\alpha]$ to evaluate a subjective formula against a $\KB$. Essentially, the function is a special case of the regression operator in~\cite{liu:ijcai} and returns a rigid formula. Thereafter, $\KB\models \alpha$ is reduced to $\models \textbf{E}[\KB,\alpha]$. For example, let $\KB$ be as Example~\ref{example:bat},
    $\textbf{E}[\KB,\bel(h=2)<1]$ returns $f(2)<1$. Since $f(2)=0$ and $\models 0 <1$, $\KB \models\bel(h=2)<1$.
    }
\end{enumerate}

\begin{figure}[t]
  \centering
   \includegraphics[width=.4\textwidth]{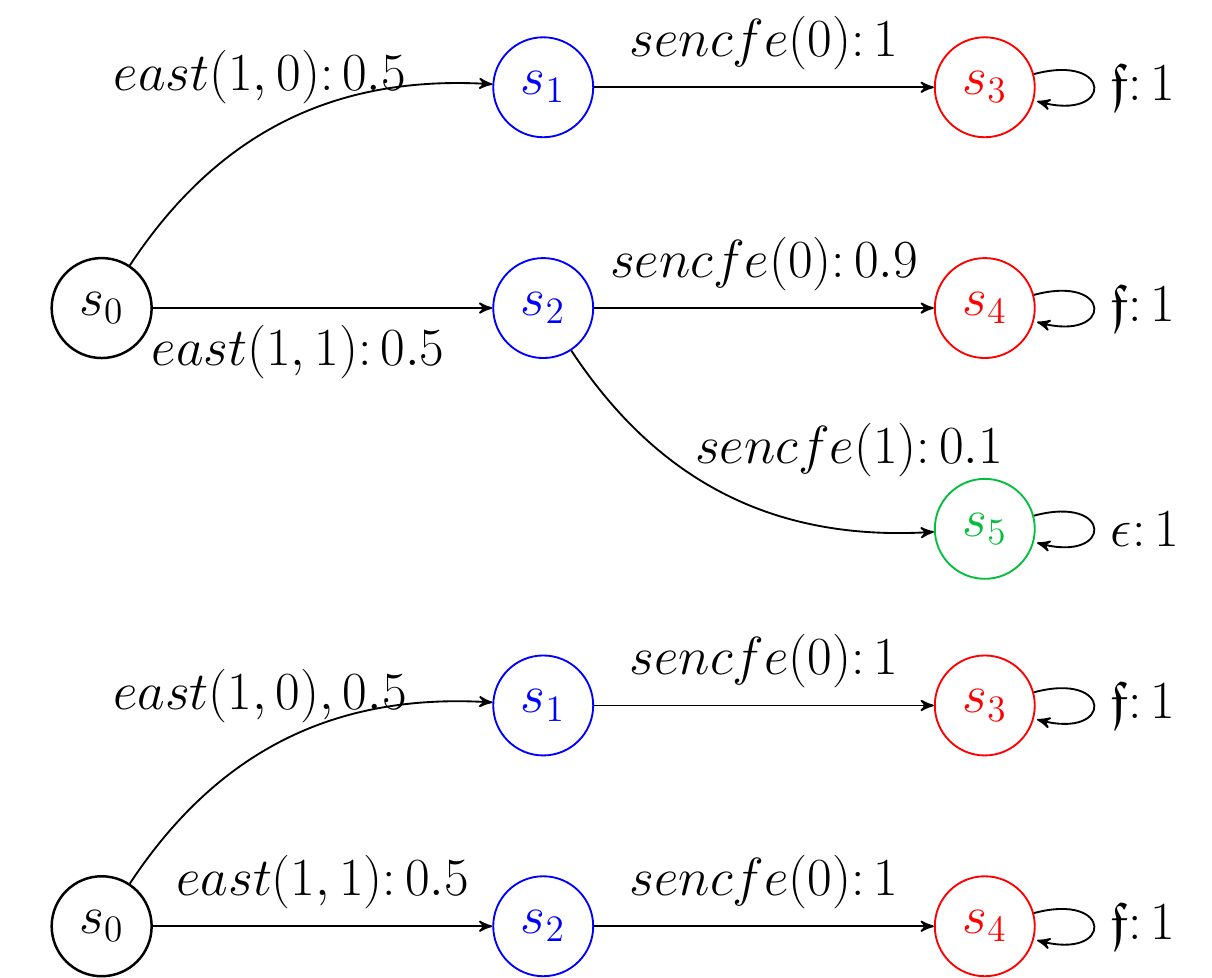}
  \caption{POMDPs induced by type $\tau_1$ (above) and  $\tau_2,\tau_3$ (below) for the coffee robot example.}
  \label{fig:pomdp}
\end{figure}

\begin{lemma}
Given a program $\calP$ and a bounded state formula $\Phi'$,
for all $e,w$ s.t. \(e,w \models \Sigma_0 \union \Sigma \union \oknow(\bel^f \land \Sigma')\),
$\mathsf{M}^{e,w}_{\pe} \models \Phi'$ iff $\mathsf{M}^{\tau}_{\pe} \models_{p} \Phi'_p$ where $\tau$ is the type of $e,w$, $\Phi'_p$ a PCTL formula obtained from $\Phi'$ by replacing all its $\lang$ sub-formula with the counter-part atomic proposition, and $\models_{p}$ is defined in the standard way \cite{norman2017verification}.
\end{lemma}
Since there are only finitely many type elements, there are only finitely many types for a given program. Hence, we can exploit existing model-checking tools like \textsc{Prism}~\cite{KNP11} or \textsc{Storm}~\cite{hensel2021probabilistic} to verify the PCTL properties against these finitely many POMDPs. Consequently, we have the following theorem.
\begin{theorem}
The verification problem is decidable for temporal properties specified by bounded state formulas.
\end{theorem}

In our coffee robot example, we obtain three types  $\tau_1$, $\tau_2$, and $\tau_3$  for worlds satisfying $\{h=0\}$, $\{h=-1\}$, and $\{h<0 \land h\neq -1\}$ in the initial state respectively. This is because $\Sigma_0$ only says $\{h\le 0\}$. The corresponding finite POMDPs are depicted in Fig.~\ref{fig:pomdp}. Note that the POMDPs for $\tau_2,\tau_3$ are the same. The observations of states are indicated by colors. Black, blue and green represent the observations of the $\KB$s with belief distribution $f$,$f'$, and $f''$ in Example~\ref{example:progression} respectively, while red stands for the observation of the $\KB$ with belief distribution $f'''$ as $f'''(u) = \textbf{if}~ u=0~\textbf{then}~ \frac{1}{3} ~\textbf{else if}~ u=1 ~\textbf{then}~ \frac{2}{3} ~\textbf{else}~ 0$.
Clearly, only the POMDP of $\tau_1$ can reach the observation $\oknow(\bel^{f''}\land \Sigma'\colon 1)$ which satisfies the label $p_{\bel(h=2 \colon 1)}$, and the probability of reaching it is $0.5 \times 0.1 =  0.05$, therefore $\calP \nvDash \mathbf{P}_{\ge 0.05}[\mathbf{F}^{\le 2} \bel(h=2 \colon 1)]$ (recall that in Def. \ref{def:verification} the satisfiability of a property for a program requires all the underlying POMDPs to satisfy the property).

\section{Related Work}
\label{sec:relatedwork}

Our formalism extends the modal logic $\lang$~\cite{liu:dsp}, a variant of \cite{belle2017reasoning}.  The idea of using the same modal logic to specify the program and its properties is inspired by the work of Cla{\ss}en and Zarrie{\ss}~\shortcite{classen2017decidable}. Similar approaches on the verification of CTL$^*$, LTL, and CTL properties of $\golog$ programs include \cite{classen2008logic,zarriess2015verification,zarriess2016decidable}. Axiomatic approaches to the verification of $\golog$ programs can be found in~\cite{de2019non,de2016verifying}.


While the verification of arbitrary $\golog$ programs is clearly undecidable due to the underlying first-order logic,
Cla{\ss}en et al. \shortcite{classen2013decidable} established decidability in case the underlying logic is restricted to the two-variable fragment, the program constructs disallow non-deterministic pick of action parameters, and the BATs are restricted to be local-effect. Later, the constraints on BATs are relaxed to acyclic and flat BATs in \cite{zarriess2016decidable}. Under similar settings, \cite{zarriess2015verification,classen2017decidable} show that the verification of $\cal{ALCOK}$-$\golog$ programs, where the underlying logic is a description logic, and $\dtgolog$ programs against LTL and PRCTL specification, respectively, is decidable. What distinguishes our work from the above is that we assume the environment is partially observable to the agent while they assume full observability.

Verifying temporal properties under partial observation has been studied extensively in model checking \cite{chatterjee2016decidable,chatterjee2016optimal,norman2017verification,bork2020verification,bork2022under}, in planning \cite{madani2003undecidability}, and in stochastic games \cite{kwiatkowska2009stochastic}. Notably the work on probabilistic planning \cite{madani2003undecidability} is closely related to our belief program verification as belief programs can be viewed as a compact representation of a plan. Moreover, it suggested that probabilistic planning is undecidable under different restrictions. Perhaps, the most relevant restriction is that probabilistic planning is undecidable even without observations, which essentially corresponds to our restriction on context-free likelihood axioms, which excludes sensing. However, our results go beyond this as we show the problem remains undecidable when restricting actions to be local-effect. Another proposal on compact representation of plans is the belief program by \cite{lang2015probabilistic}. Nevertheless, the proposal is primitive as the underlying logic is propositional, i.e., beliefs are only about propositions. Hence, 
verification there reduces to regular model-checking. In contrast, our framework based on the logic $\lang$ which allows us to express incompleteness about the underlying model. Therefore, to verify a belief program, one has to perform model-checking for potentially infinitely many POMDPs. Other virtues of our belief program, to name but a few, include that 1) tests of the program can refer to beliefs about belief, i.e. meta-beliefs, and beliefs with quantifying-in 2) we can express that dynamics of a domain that holds in the real world are different from what the agent believes (Our coffee robot is an example of this kind). Hence, although \cite{lang2015probabilistic} showed that the verification problem is decidable when restricting to finite horizon, our result on decidability goes beyond them since our problem is more general than theirs.

\section{Conclusion}
\label{sec:conclusion}

We reconsider the proposal of belief programs by Belle and Levesque based on the logic $\lang$.
Our new formalism allows, amongst others, to 
define the transition system and specify the temporal properties like \emph{eventually} and \emph{globally}
more smoothly. Besides, we study the complexity of the verification problem. As it turns out, the problem is undecidable even in very restrictive settings. We also show a case where the problem is decidable. 

As for future work, there are two promising directions. On the complexity of verification, whether it is decidable or not remains open for the case where the SSAs and LA are context-free. Our sense is that, under such a setting, belief programs in general cannot simulate arbitrary probabilistic automata, but only a subset. Since the emptiness problem of probabilistic automata is a special case of the verification problem, evidence showing undecidability of  emptiness problem for such a subset could prove the undecidability for the verification problem for programs with context-free SSAs and LA.
Besides, \cite{chatterjee2012decidable,fijalkow2012deciding} show a set of decidable decision problems in related to special types of probabilistic automata. It is interesting to see how these problem can be transformed to the verification problem and hence find decidable cases.
Another direction is more practical. It is desirable to design a general algorithm to perform verification of arbitrary belief programs, even if the algorithm might not terminate. In this regard, symbolic approaches in solving first-order MDP and first-order POMDP \cite{sanner2009practical,sanner2010symbolic}, compact representations of (infinite) (PO)MDPs, are relevant.

\section*{Acknowledgments}
This work has been supported by the Deutsche Forschungsgemeinschaft (DFG, German Research Foundation) RTG 2236 ‘UnRAVeL’ and by the EU ICT-48 2020 project TAILOR (No. 952215). Special thanks to the reviewers for their invaluable comments.

\bibliographystyle{kr}
\bibliography{kr}

\end{document}